# Deeper Learning with CoLU Activation Function

**Advait Vagerwal**

advaitv0709@gmail.com

## Abstract

In neural networks, non-linearity is introduced by activation functions. One commonly used activation function is Rectified Linear Unit (ReLU). ReLU has been a popular choice as an activation but has flaws. State-of-the-art functions like Swish and Mish are now gaining attention as a better choice as they combat many flaws presented by other activation functions. CoLU is an activation function similar to Swish and Mish in properties. It is defined as f(x)=x/(1-x^-(x+e^x)). It is smooth, continuously differentiable, unbounded above, bounded below, non-saturating, and non-monotonic. Based on experiments done with CoLU with different activation functions, it is observed that CoLU usually performs better than other functions on deeper neural networks. While training different neural networks on MNIST on an incrementally increasing number of convolutional layers, CoLU retained the highest accuracy for more layers. On a smaller network with 8 convolutional layers, CoLU had the highest mean accuracy, closely followed by ReLU. On VGG-13 trained on Fashion-MNIST, CoLU had a 4.20% higher accuracy than Mish and 3.31% higher accuracy than ReLU. On ResNet-9 trained on Cifar-10, CoLU had 0.05% higher accuracy than Swish, 0.09% higher accuracy than Mish, and 0.29% higher accuracy than ReLU. It is observed that activation functions may behave better than other activation functions based on different factors including the number of layers, types of layers, number of parameters, learning rate, optimizer, etc. Further research can be done on these factors and activation functions for more optimal activation functions and more knowledge on their behavior.

## 1 Introduction

Activation functions are used in machine learning to introduce non-linearity in neural networks. These functions play an important role in the training process in a neural network. Rectified Linear Unit (ReLU) [1] is a commonly used activation function. ReLU was created keeping certain properties in mind. While ReLU is popularly used because of its simplicity and effectiveness, it has a few undesirable flaws. These include exploding gradient [20], non-differentiability, positive mean activation [5], dying ReLU [21], etc. Many activation functions, like Exponential linear unit (ELU) [5], Parametric Rectified Linear Unit (PReLU) [6], Scaled Exponential Linear Unit (SELU) [7], etc., have been proposed as a substitute for ReLU. Recently, two functions Mish [4] and Swish [2] have been introduced and consistently outperform other activation functions.

In this paper, Collapsing Linear Unit, or CoLU in short, is introduced. CoLU is defined as f(x)=x/(1-x^-(x+e^x)). Based on experiments, this function outperforms ReLU, Swish, and Mish on neural networks trained on MNIST, Fashion-MNIST, and Cifar-10 datasets. On MNIST, CoLU performs better than SELU, ReLU, Mish, and Swish on neural networks with an incrementally increasing number of convolutional layers. On a smaller network with 8 convolutional layers, CoLU had the highest mean accuracy, closely followed by ReLU. On VGG-13 trained on Fashion-MNIST, CoLU had a 4.20% higher accuracy than Mish and 3.31% higher accuracy than ReLU. On ResNet-9 trained on Cifar-10, CoLU had 0.05% higher accuracy than Swish, 0.09% higher accuracy than Mish and 0.29% higher accuracy than ReLU.

## 2 Collapsing Linear Unit (CoLU)

Collapsing Linear Unit, or CoLU is defined as:

$$f(x) = \frac{x}{1 - xe^{-(x+e^x)}}$$

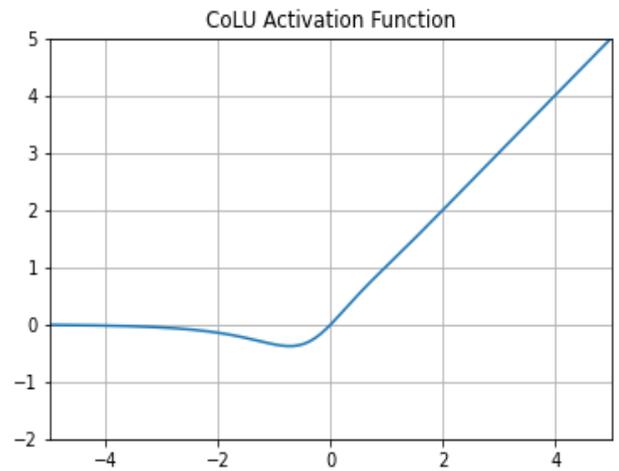

CoLU is similar to Mish and Swish and has similar properties to them. CoLU has a range of [≈-0.3762, ∞). Mish has a range of [≈-0.3087, ∞), while Swish has a range of [≈-0.2784, ∞). As compared to Mish and Swish, CoLU has a bigger "bump". The derivative of CoLU is defined as:

$$f'(x) = \frac{\lambda(\lambda - x^2(e^x + 1))}{(\lambda - x)^2} \quad \text{Where } \lambda = e^{(e^x + x)}$$

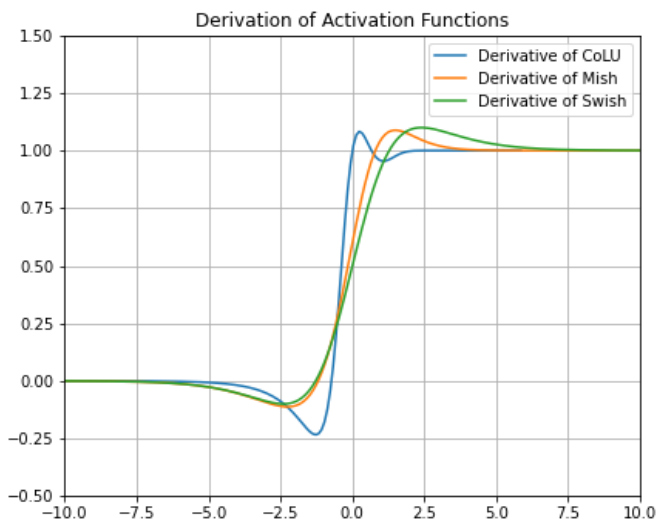
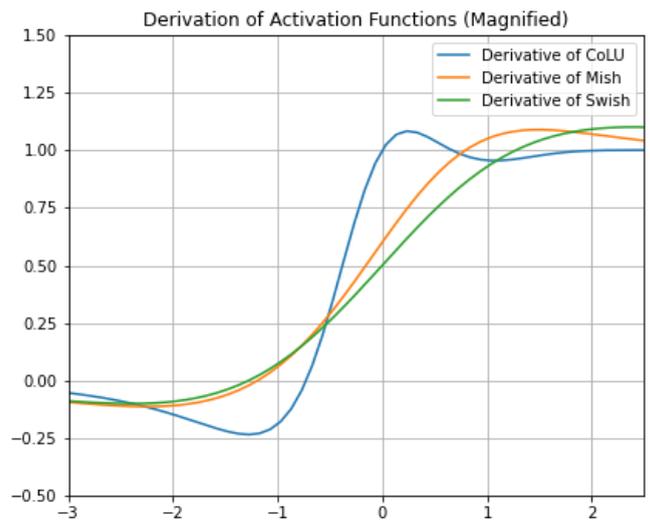

Since this function is unbounded above, it avoids saturation. Saturation of outputs can negatively affect training and may slow down the training process due to near-zero gradient [13]. Being bounded below helps in regularizing the outputs. Similar to both Swish and Mish, CoLU is smooth and non-monotonic. The smoothness of activation functions is desirable as the continuous derivative of the function can be calculated. The non-monotonic property helps a few negative values to be preserved, which helps during back-propagation in the neural network.

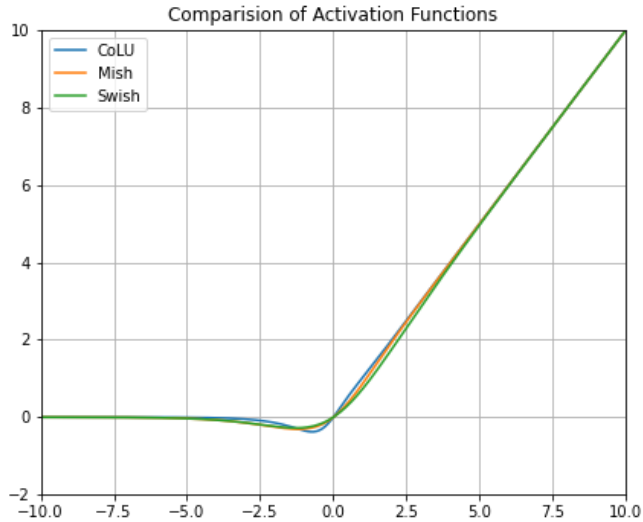 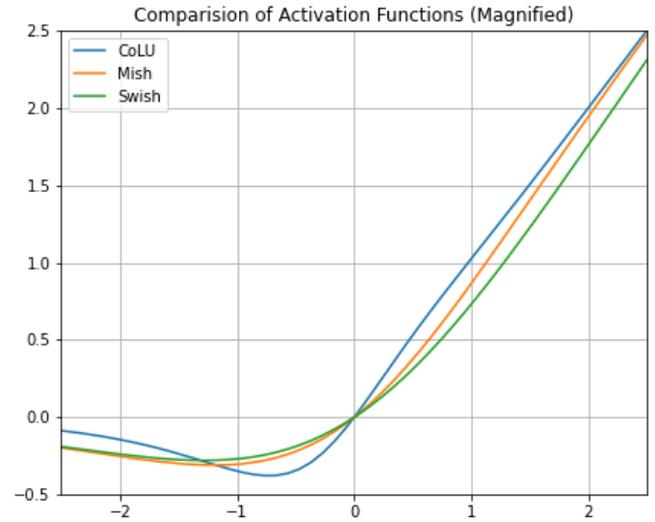

CoLU can be defined in Tensorflow [11] as x/(1-x*(tf.math.exp(-x-tf.math.exp(x)))). In Torch [12] CoLU is defined as x/(1-x*(torch.exp(-x-torch.exp(x)))). It is observed CoLU usually performs better with Batch normalization layers after convolution and fully-connected layers. A low learning rate is recommended.

| Function | Mathematical Formula | Properties |
| --- | --- | --- |
| CoLU | $\dfrac{x}{1-xe^{-(x+e^x)}}$ | Continuously differentiable, unbounded above, bounded below, non-monotonic, non-saturated above, computationally expensive |
| ReLU [1] | $\begin{cases} x & if\ x \geq 0 \\ 0 & if\ x < 0 \end{cases}$ | Non-differentiable at 0, unbounded above, bounded below, monotonic, non-saturated above, computationally cheap |
| Swish [2] | $\dfrac{x}{1+e^{-x}}$ | Continuously differentiable, unbounded above, bounded below, non-monotonic, non-saturated above, computationally expensive |
| Logistic Sigmoid [3] | $\dfrac{1}{1+e^{-x}}$ | Continuously differentiable, bounded below and above, monotonic, saturated above, computationally expensive |
| Mish [4] | $\dfrac{x((1+e^x)^2-1)}{((1+e^x)^2+1)}$ | Continuously differentiable, unbounded above, bounded below, non-monotonic, non-saturated above, computationally expensive |
| ELU [5] | $\begin{cases} x & if\ x \geq 0 \\ \alpha(e^x-1) & if\ x < 0 \end{cases}$ | Non-differentiable at 0, unbounded above, bounded below, monotonic, non-saturated above, computationally expensive, uses a parameter |

| TanH [3] | $\dfrac{e^x - e^{-x}}{e^x + e^{-x}}$ | Continuously differentiable, bounded below and above, monotonic, saturated above, computationally expensive |
| --- | --- | --- |
| Softplus [10] | $ln(1 + e^x)$ | Continuously differentiable, unbounded above, bounded below, monotonic, non-saturated above, computationally expensive |

# 3 Experiments

## 3.1 MNIST

To observe how the training was affected by the increase in the total number of convolutional layers, neural networks with increasing numbers of convolutional layers with varying depths were trained with batch normalization [14] and max-pooling layers [19] after each convolution layer with a dropout [15] rate of 25%. Every convolutional layer had an L2 regularizer with regularizer factor 1e-4. Stochastic Gradient Descent optimizer [16] was used with a learning rate of 0.001 [17], decay of 1e-4 [17], and momentum of 0.9 [17,18]. The models were trained on a batch size of 64 on 30 epochs. The results are summarized by the following graph. In this experiment, CoLU was compared with SELU, Mish, Swish, and ReLU.

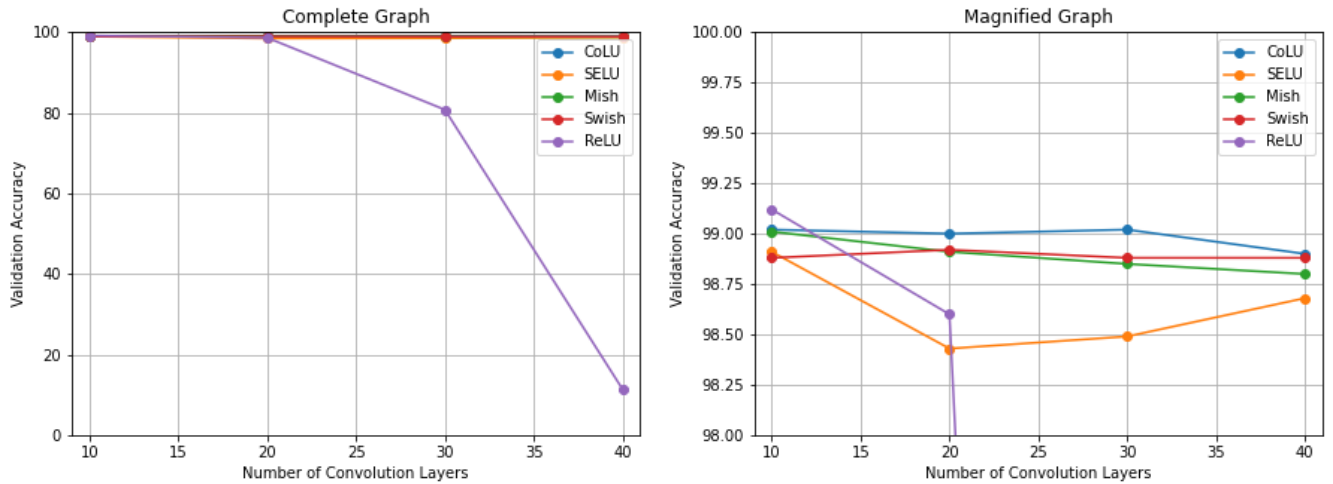

It is observed that ReLU outputted the highest accuracy on 10 layers, followed by CoLU and Mish. While the accuracy dropped substantially for ReLU for an increasing number of convolution layers, CoLU maintained the highest accuracy and peaked at 30 layers. From this experiment, it seems that ReLU still outputs the highest accuracy on small neural networks. So, a smaller neural network with 8 convolutional layers and a similar structure was trained. The results are summarised in the following table. For the most accurate results, the same model was trained 10 times and the mean value of accuracy and loss was calculated with their respective standard deviations.

| Function | Mean accuracy | Std. deviation of accuracy | Mean cross-entropy loss | Std. deviation of loss |
|---|---|---|---|---|
| CoLU | 99.290% | 0.093 | 0.10172 | 0.00298 |
| ReLU | 99.288% | 0.099 | 0.10133 | 0.00302 |
| Swish | 99.248% | 0.117 | 0.10056 | 0.00385 |
| Mish | 99.270% | 0.106 | 0.10116 | 0.00402 |

In this experiment, CoLU outperformed other functions closely followed by ReLU. This shows that CoLU can work better than other activations on smaller neural networks as well as deeper neural networks.

## 3.2 Fashion-MNIST

For testing this function on the Fashion-MNIST dataset, a VGG-13 model was trained with a batch size of 64 on 100 epochs. SGD optimizer with a learning rate of 0.001, decay of 1e-4, and momentum of 0.9 was used. The neural network was trained 5 times for all functions. The mean testing accuracies and losses of CoLU, ReLU, Mish, and ELU (α=1) were measured:

| Function | Accuracy | Cross-entropy loss |
|---|---|---|
| CoLU | 92.15% | 0.4261 |
| ReLU | 88.84% | 0.8915 |
| Mish | 87.95% | 0.6594 |
| ELU (α=1) | 90.38% | 0.5422 |

CoLU had the highest accuracy and resulted in 4.20% higher accuracy than Mish and 3.31% higher accuracy than ReLU. This again shows that CoLU does significantly well on deeper neural networks as compared to other activation functions. Note that while Mish consistently outperforms most other activations, it resulted in the lowest mean accuracy as compared to CoLU, ReLU, and ELU.

## 3.3 Cifar-10

On Cifar-10, a Resnet-9 model [8] was trained to compare different activation functions with CoLU. A batch size of 400 was used on 100 epochs. SGD optimizer with a learning rate of 0.001, decay of 1e-4, and momentum of 0.9 was used and data augmentation was applied. The average accuracy of these functions on 5 runs was recorded and shown below:

| Function | Accuracy | Cross-entropy loss |
|---|---|---|
| CoLU | 88.51% | 0.427 |
| ReLU | 88.22% | 0.427 |
| Mish | 88.42% | 0.415 |
| Swish | 88.46% | 0.425 |

The overall mean accuracy of CoLU was higher than other functions in this experiment, but the mean loss was higher than Mish and Swish. The higher accuracy of CoLU activation shows the consistency of this function over other functions. With proper parameters and layers, CoLU can outperform other activations on almost every neural network and data.

Another noteworthy observation of these experiments is that different activation functions may perform better than others based on different factors like the number of layers, types of layers, number of parameters, learning rate, optimizer, etc. As observed in experiment 3.1, ReLU outperformed other activations on 10 layers followed by CoLU and Mish, while on 40 layers CoLU had the highest accuracy followed by Swish and Mish. When a neural network with 8 convolutional layers was trained, CoLU had the highest accuracy closely followed by ReLU. In experiment 3.2, using Mish as an activation resulted in the lowest accuracy as compared to other activations in the experiment. In experiment 3.3, Swish had a higher mean accuracy than Mish.
In short, different activation functions may result in better accuracies than other activation functions, depending on the factors in training.

# 4 Conclusion and Future Work

Collapsing Linear Unit, or CoLU is a new activation function defined as $f(x) = x/(1 - xe^{-(x+e^x)})$. This activation function is similar to Swish and Mish and has desirable qualities. This function outperforms other activation functions on many different neural networks trained on different data. In general, it is observed that CoLU performs significantly better than other functions on deeper models like VGG-13. Based on the experiments performed in this paper, different activations may perform better than other functions based on different factors like the number of layers, types of layers, number of parameters, learning rate, optimizer, etc. of the neural network. Looking forward, further research can be done on the properties of activation functions and why one function may perform better than other functions in particular cases. New activation functions with more desirable properties can be discovered as a further boost in training accuracy.